\documentclass[wcp]{jmlr}
\usepackage{longtable}
\usepackage{amsmath}

\usepackage{booktabs}

\title[Evaluating a Simple Retraining Strategy as a Defense Against Adversarial Attacks]{Evaluating a Simple Retraining Strategy as a Defense Against Adversarial Attacks}

  \author{\Name{Nupur Thakur} \Email{nsthaku1@asu.edu}\\
 \Name{Yuzhen Ding} \Email{yding58@asu.edu}\\
  \Name{Baoxin Li} \Email{Baoxin.Li@asu.edu}\\
  \addr Arizona State University}

\begin{document}

\maketitle

\begin{abstract}
Though deep neural networks (DNNs) have shown superiority over other techniques in major fields like computer vision, natural language processing, robotics, recently, it has been proven that they are vulnerable to adversarial attacks. The addition of a simple, small and almost invisible perturbation to the original input image can be used to fool DNNs into making wrong decisions. With more attack algorithms being designed, a need for defending the neural networks from such attacks arises. Retraining the network with adversarial images is one of the simplest techniques. In this paper, we evaluate the effectiveness of such a retraining strategy in defending against adversarial attacks. We also show how simple algorithms like KNN can be used to determine the labels of the adversarial images needed for retraining. We present the results on two standard datasets namely, CIFAR-10 and TinyImageNet.
\end{abstract}

\section{Introduction}

Deep Neural Networks have shown excellence in many tasks like image classification, image segmentation, image reconstruction etc. Though they have outperformed humans in some of these tasks, they are highly vulnerable to adversarial attacks. A simple change in the input image can change the output of the network drastically, which is highly undesirable. 

There are several ways to categorize the adversarial attacks. First, they can be classified as targeted attacks and non-targeted attacks. In a targeted attack, the goal is to misguide the network to a specific targeted class label different than the true label. On the other hand, in a non-targeted attack, the goal is to force the network to predict any label different than the original label. The targeted attacks are more difficult to design than the non-targeted ones, as the attack is considered successful only when the network predicts the specified category.

The adversarial attacks can also be divided as a white-box attack or black-box attack, depending on the extent of access the attacker has to the architecture of the targeted network. In white-box attack, the attacker has access to the entire architecture of the targeted network. This makes the design of such attacks simpler than the black-box attacks because the architecture of the target network is completely unknown to the black-box attackers. Having full access to the architecture of the targeted network is a strong assumption which may be unrealistic in many real-world applications and thus black-box attacks have drawn more attention recently. Apart from the attacks, the perturbations added to the input image are of two types - universal perturbations and image-dependent perturbations.  Universal perturbations are fixed patterns which can be added to any clean image in order to generate adversarial images. On the contrary, image-dependent perturbations vary from image to image. 

Neural networks are being used in applications like surveillance and user authentication, where misclassification due to adversarial attacks can pose major security threats. In applications like autonomous driving (\cite{sermanet2011traffic}), misclassification can lead to fatal accidents. Therefore, defending the neural networks against adversarial attacks is important. A basic way to defend against the adversarial attacks is to retrain the network with adversarial examples (\cite{szegedy2013intriguing}). In this paper, we present an analysis of how retraining the network with adversarial images can be an effective defense against the adversarial attacks. We also show how simple machine learning algorithms may be used to determine the labels of the adversarial images, which are necessary for retraining. The rest of the paper is organised as follows: Section 2 includes important related work. Section 3 explains the retraining strategy and representative attackers. Section 4 includes our experiments and Section 5 concludes.

\section{Related Work}

In this section, we present a brief survey of the recent work related to adversarial attacks and the defenses. The phenomenon that the network gets fooled by a small change in input image was initially studied by \cite{szegedy2013intriguing}. The added perturbations are so small that the images are recognizable by the human eye but the network fails to categorize it correctly.

\cite{nguyen2015deep} showed that it takes little effort to generate random noise images for which the network predicts a label with extremely high confidence. The images look like noise to the human eye but the network assigns them a class label with as high confidence as 99\%. The robustness of the adversarial images is demonstrated in \cite{athalye2017synthesizing}. They proposed a new method called Expectation over Transformation to generate adversarial images which are highly robust to different transformations like re-scaling, rotation, the addition of Gaussian Noise, lightening or darkening by an additive factor. \cite{eykholt2018robust} presents an adversarial attack known as Robust Physical Perturbation $(RP_2)$ that can create perturbations under physical conditions. 

When it comes to defenses against these attacks, retraining the network with the adversarial images was discussed in \cite{szegedy2013intriguing}. The generated adversarial images were used for training the network. The error rate on such images was seen to decrease drastically after retraining the network. It states that retraining is different from data augmentation techniques since such type of data does not occur frequently in the test set. \cite{papernot2016distillation} introduced another defense strategy known as defensive distillation to defend adversarial images. It is based on the distillation mechanism which was introduced in \cite{hinton2015distilling} to reduce the computational complexity of the large neural network architectures. 

\section{Defending Attacks via a Simple Retraining Strategy}

Given a classifier trained on some original training set of images, retraining is the process of training the classifier again with adversarial images added to the original training set in order to increase its robustness. Generally speaking, adversarial images do not come with labels that can be directly used for retraining. As an adversarial image typically looks similar to certain original image, the ``true" label of this adversarial image should be the label of that original image (although by definition of ``attacking", the classifier may not be able to predict that label). When the adversarial images do not look like any original images but more like random noise, they are known as fooling images. In such cases, it is not proper to assign any original label to them, and one way is to put all the fooling images in one new class (or possibly multiple $n$ new classes depending on some clustering analysis of the images). Therefore, there may be an addition of new class labels while retraining with fooling images.

\subsection{Producing True Labels for the Adversarial Images}

Attacking algorithms often start from an authentic image (of the certain label) to produce an adversarial or fooling image. Hence, in theory, the original label may be used as the true label of the generated image. However, having access to the attacker in order to determine the true labels is not always possible. Also, it has been demonstrated that adversarial images are transferable. This means that the adversarial images can fool several neural networks of different architecture simultaneously. Therefore, it may be problematic to rely on a single neural network to attempt to predict the true label of an adversarial image. 

\begin{figure*}[h]
    \begin{center}
       \includegraphics[scale = 0.3]{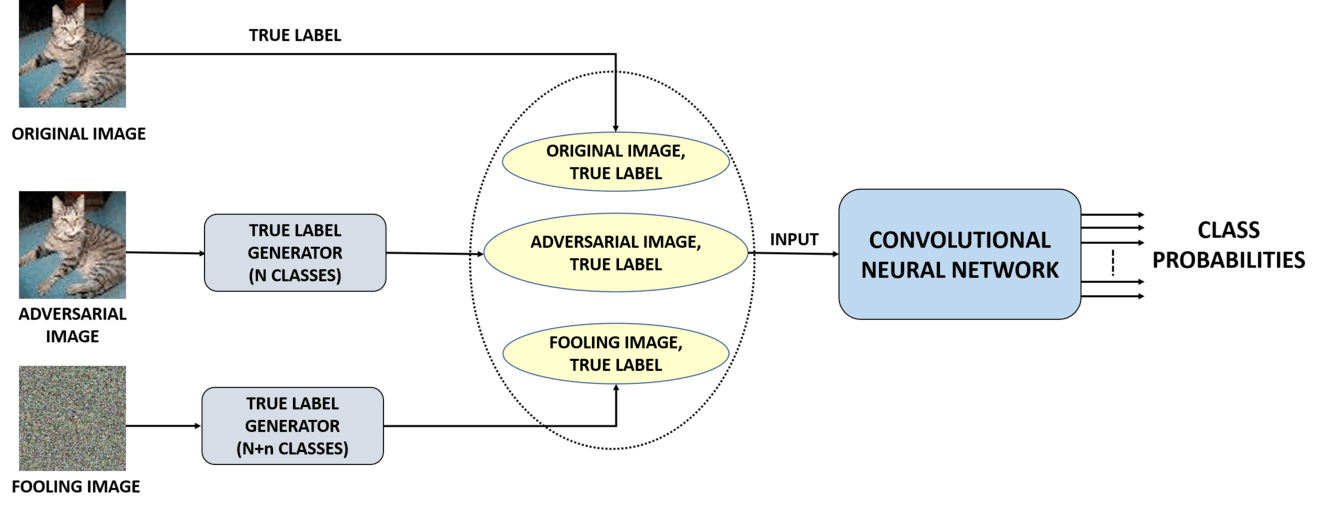}
    \end{center}
\caption{A general framework for the simple retraining strategy. The true label generator determines the labels of adversarial and fooling images using network-independent techniques like KNN. The images with their true labels are then used for retraining the network.}
    \label{retraining}
\end{figure*}

For the above considerations, we should resort to some universal technique not specific to any particular network. In this study, we employ the K-Nearest Neighbor (KNN) algorithm for this purpose. Most of the adversarial attacks have a constraint of maintaining a minimum $L_2$ distance between a generated adversarial image and an original image to make them look similar to each other. Due to this constraint, the KNN algorithm can be used as a simple strategy for finding out the true labels of the adversarial images. For fooling images, the $L_2$ distance between them and original images is very large. Hence while using the KNN algorithm for fooling images, we may create one additional class if the distance becomes too large (alternatively, multiple new classes may be created to accommodate clusters of the fooling images). 

When the number of dimensions is high, KNN takes a longer time to run. Higher the dimensions, longer is the running time and higher is the accuracy. In order to reduce dimensions to lessen the running time, PCA (Principal Component Analysis) may be used. The number of reduced dimensions need to be chosen such that high accuracy is achieved in a reasonable amount of time. Figure~\ref{retraining} shows the complete framework for the retraining strategy with adversarial and/or fooling images. 

\subsection{Choosing the Representative Attackers}

We selected four adversarial attacks to evaluate the effect of retraining as a defense strategy which are briefly discussed here. There are two main reasons for choosing these attacks. First, they produce strong adversarial images that lead to high initial fooling ratio. All of these recent attacks have different working principles. For instance, Generative Adversarial Perturbations attack makes use of a neural network to create the perturbations, FGSM is gradient-based, DeepFool attack is about linearizing the classifier to generate perturbations and Carlini-Wagner attack is optimization-based. Hence, these approaches represent a wide range of possibilities that may be employed by an attacker.

\subsubsection{Generative Adversarial Perturbations (GAP)}

\begin{figure*}[h]
    \begin{center}
           \includegraphics[scale = 0.3]{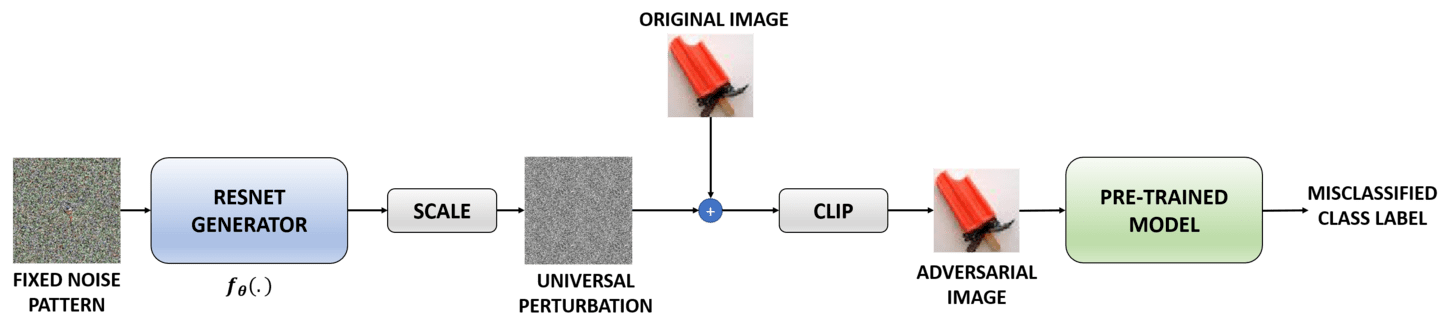}
    \end{center}
\caption{The framework for generating universal perturbations. The ResNet receives a noisy pattern as input and produces a universal perturbation pattern. This can be added to an input image to convert it to an adversarial image, which is then fed to the network as input.}
\label{CVPR2018}
\end{figure*}

The framework used for generating the universal perturbations in this technique (\cite{poursaeed2018generative}) is shown in Figure~\ref{CVPR2018}. A neural network $f_\theta (.)$ is used in order to approximate the function that can generate the perturbations for converting the original images into adversarial ones. Here, $f_\theta (.)$ is a ResNet, which is generally used for transforming images from one domain to another. 

To generate the universal perturbations, a fixed pattern drawn from a uniform distribution randomly is passed to the generator as input. The output, again a fixed pattern, is then scaled to the range on which the network is trained. This is the final universal perturbation which can be added to any natural image from the dataset to make it an adversarial image. For the rest of the paper, we refer this method as GAP attack. 

\subsubsection{DeepFool}

DeepFool (\cite{moosavi2016deepfool}) is a non-targeted, white box attack which is based on iterative linearization of the target classifier. It produces minimum perturbations, imperceptible to the human eye, just enough to change the prediction of the classifier. It is assumed that the neural networks are completely linear and the hyperplanes separate the classes from one another. Using this assumption, a greedy approach and existing optimization techniques, the minimally perturbed image is found. It uses the $L_2$ distance metric.

The minimum perturbations reduce the complexity and the time taken to generate the adversarial images. Therefore, it is a highly efficient method and produces closer adversarial examples. 

\subsubsection{Carlini-Wagner Attack}

This is an optimization-based (\cite{carlini2017towards}), white-box attack that uses the perturbation norm in order to design the cost function used for producing the adversarial images. Similar to L-BFGS method (\cite{szegedy2013intriguing}), this algorithm makes use of box constraints while defining the cost function. The Carlini-Wagner algorithm covers three different attack algorithms based on $L_0$, $L_1$ and $L_2$ distance metric. These attacks have a success rate of almost 100\%. The $L_2$ attack is the strongest one among all the three attacks. 

The invisible perturbations in the adversarial images and high success rate of fooling make it a very strong adversarial attack. However, the algorithm has high computational cost and time complexity for adversarial image generation.

\subsubsection{Fast Gradient Sign Method (FGSM)}

\begin{figure}[h]
    \begin{center}
    \label {framework}
       \includegraphics[scale = 0.5]{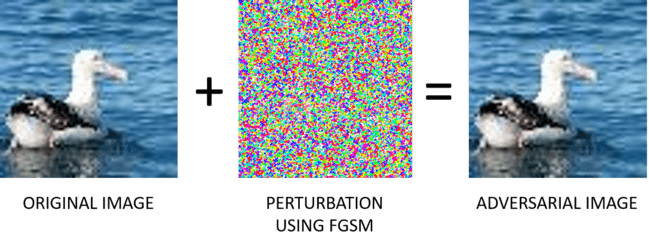}
    \end{center}
\caption{Fast Gradient Sign Method (FGSM). The final adversarial image is generated by adding the perturbation calculated using FGSM method to the original image from the TinyImageNet dataset.}
\label{fgsm}
\end{figure}

Fast Gradient Sign Method (FGSM) (\cite{goodfellow2014explaining}) is one of the initial adversarial attacks. It makes use of the gradients of the network in order to produce perturbed images which are misclassified by the network. It is a non-targeted, white box attack. If $\theta$ is the parameters of the neural network, $x$ and $y$ are the input and corresponding target respectively, then the adversarial image $x^*$ is calculated using Equation~\eqref{fgsmeq}. Figure~\ref{fgsm} shows how the perturbations added to the original image are invisible to the human eye. 

\begin{equation}
\label{fgsmeq}
    x^* = x + \epsilon\text{sign}\left({\nabla_x}{J\left(\theta, x, y\right)}\right)
\end{equation}

\section{Experiments and Results}

In this section, our experiments conducted on CIFAR-10 and TinyImageNet dataset for evaluation of retraining strategy as a defense against adversarial attacks are included. CIFAR-10 dataset (\cite{krizhevsky2014cifar}) contains 50,000 training images and 10,000 test images of size 32x32. TinyImageNet dataset is the smaller version of ImageNet dataset (\cite{deng2009imagenet}) containing 200 classes. Each class of the dataset contains 500 training images, 50 test images and 50 validation images of size 64x64. We use the validation data of TinyImageNet for our experiments.

For CIFAR-10 dataset, VGG16 (\cite{simonyan2014very}) network with a test accuracy of 92.37\% is used as the target classifier. In the case of TinyImageNet dataset, a ResNet (\cite{he2016deep}) classifier is used which achieves an accuracy of 72.42\% on the validation data.

\subsection{Black-Box Attacks}
\subsubsection{Generative Adversarial Perturbations}
We consider non-targeted attacks using ResNetGenerator for generating universal perturbations. The perturbations are controlled by L-infinity which is 10 for both the datasets.

\begin{figure}[h]
    \begin{center}
       \includegraphics[scale = 1]{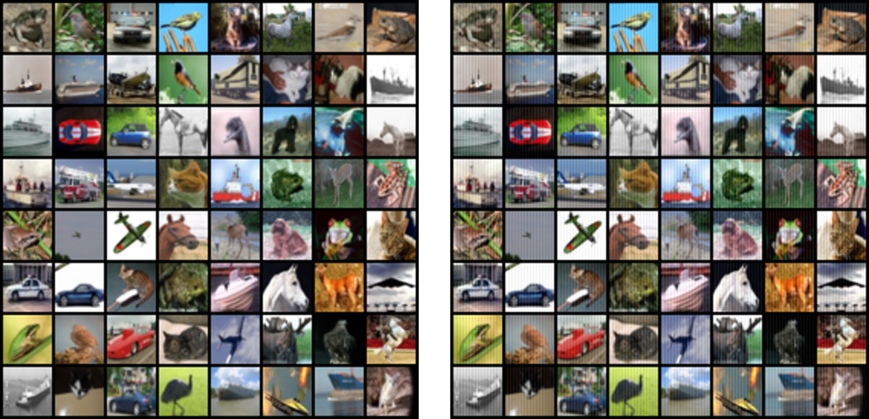}
    \end{center}
\caption{Original images and adversarial images generated using the GAP attack for CIFAR-10 Dataset. Images on the left are the original images and the corresponding adversarial images are displayed on the right. The added perturbations are clearly visible.}
\label{cvprcifar}
\end{figure}

For CIFAR-10 dataset, the ResNetGenerator is trained with a learning rate of 0.002. The initial fooling ratio is 82\% without affecting the test accuracy on original images. Figure~\ref{cvprcifar} shows some original images and the adversarial images generated by adding universal perturbations to them. It is evident that though the images look similar to original images, the perturbations are visible to the human eye. Next, the classifier is trained with different number of adversarial images added to the original training dataset. Table~\ref{cifar10} shows the results of retraining the classifier and the fooling ratio after it. 

\begin{table*}[h]
  \centering
  \begin{tabular}{p{2.8cm}p{2.2cm}p{2.4cm}p{3.1cm}}
  \toprule
    \bf Adversarial Images used for Retraining     & \bf Accuracy on Original Images     & \bf Accuracy on Adversarial Images  &\bf Fooling Ratio \hspace{1.9cm} after Retraining  \\
    \midrule
    \hspace{0.9cm}50,000   & \hspace{0.3cm}92.25\%  &\hspace{0.4cm}91.09\%    &\hspace{0.9cm}13\%     \\
    \midrule
    \hspace{0.9cm}25,000     &\hspace{0.35cm}92.4\%  &\hspace{0.4cm}90.18\%   &\hspace{0.9cm}16\%      \\
    \midrule
    \hspace{0.9cm}12,500     &\hspace{0.35cm}91.5\%   &\hspace{0.45cm}88.5\%  &\hspace{0.9cm}24\% \\
    \midrule
    \hspace{1cm}9,000    &\hspace{0.3cm}90.52\% &\hspace{0.4cm}86.64\%   &\hspace{0.9cm}26\%    \\
    \bottomrule
  \end{tabular}
  \caption{Retraining the CIFAR-10 classifier with adversarial images generated using the GAP attack. The number of original training images is 50,000 which remains constant and the number of adversarial images used for training is varied.}
    \label{cifar10}
\end{table*}

The accuracy on the adversarial images from the Table~\ref{cifar10} shows that the network can learn the adversarial images very well. After retraining, the fooling ratio drops significantly. Even if the number of adversarial images is decreased to almost $(1/6)^{th}$ of the total training data, the network learns them very well and can no longer be fooled. 

\begin{figure}[h]
    \begin{center}
       \includegraphics[scale = 0.45]{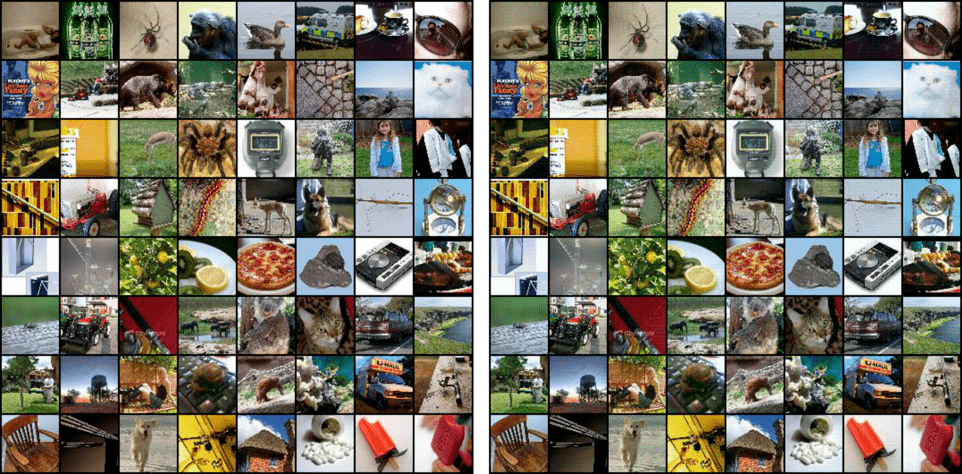}
    \end{center}
\caption{Original images and adversarial images generated using the GAP Attack for TinyImageNet dataset. Images on the left are the original images and the corresponding adversarial images are displayed on the right.}
\label{cvprtiny}
\end{figure}

For TinyImageNet, the ResNetGenerator is trained with a learning rate of 0.001. The initial fooling ratio is 92\%. As the accuracy of the classifier on the validation data is not very high, it is easier to fool the network. Therefore, the fooling ratio is higher than the accuracy of the network. Figure~\ref{cvprtiny} shows the original images and the generated adversarial images for TinyImageNet dataset. Table~\ref{tinyimagenet} shows the results of retraining the TinyImageNet classifier and re-attacking it.

\begin{table*}[h]
  \centering
  \begin{tabular}{p{2.8cm}p{2.2cm}p{2.4cm}p{3.1cm}}
\toprule
    \bf Adversarial Images used for Retraining     & \bf Accuracy on Original Images     & \bf Accuracy on Adversarial Images  &\bf Fooling Ratio after Retraining  \\
    \midrule
     \hspace{0.8cm}100,000   &\hspace{0.3cm}72.4\%  &\hspace{0.4cm}66\%    &\hspace{0.9cm}22\%     \\
\midrule
     \hspace{0.9cm}75,000     &\hspace{0.35cm}72\%  &\hspace{0.4cm}66\%   &\hspace{0.9cm}23\%       \\
\midrule
     \hspace{0.9cm}50,000     &\hspace{0.3cm}72.1\%   &\hspace{0.4cm}66\%  &\hspace{0.9cm}22\% \\
\midrule
     \hspace{0.9cm}25,000     &\hspace{0.35cm}72\%  &\hspace{0.4cm}65\%    &\hspace{0.9cm}24\%     \\
\bottomrule
  \end{tabular}
  \caption{Retraining the TinyImageNet classifier with different number of adversarial images generated using the GAP Attack. The number of original training images is 100,000 which remains constant and the number of adversarial images used for training is varied.}
    \label{tinyimagenet}
\end{table*}

The results are similar to the results of CIFAR-10 dataset. The accuracy on the adversarial images is as high as the test accuracy on original images. Furthermore, it can recognize such adversarial data after retraining indicated by low fooling ratio. This trend is observed even when only 25,000 adversarial images are used for retraining. For TinyImageNet, even $(1/4)^{th}$ of the total data is sufficient for the network to be robust to such adversarial images. Therefore, the adversarial images generated using GAP attack are not very strong as they can be easily defended by retraining the classifier with a small number of adversarial images.

\subsection{White-Box Attacks}

\subsubsection{DeepFool}

For CIFAR-10 dataset, this non-targeted attack has an initial fooling ratio of 99\%. After adding 50,000 adversarial images to original training data and 10,000 adversarial images to original test data, the test accuracy on adversarial images is 90.13\%. When this retrained classifier is attacked again using DeepFool algorithm, the fooling ratio is 99\% again. 

\begin{figure*}[h]
    \begin{center}
       \includegraphics[scale = 1]{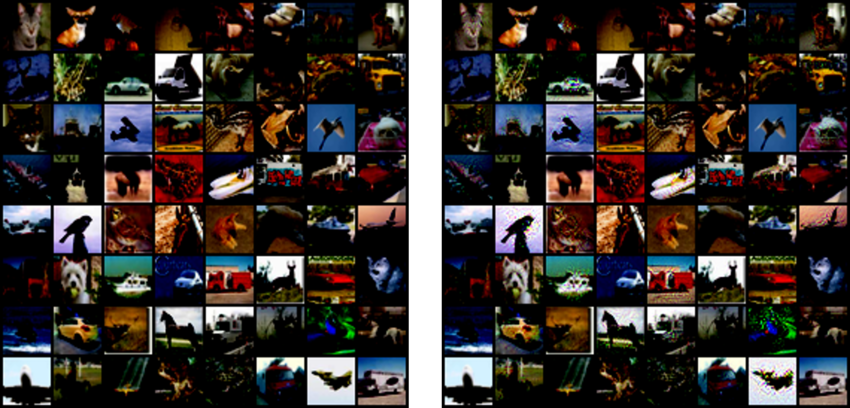}
    \end{center}
\caption{Original images and adversarial images generated using DeepFool algorithm for CIFAR-10 Dataset. Images on the left are the original images and the corresponding adversarial images are displayed on the right.}
    \label{cifar10deep}
\end{figure*}

For TinyImageNet dataset too, the initial fooling ratio is 99\%. After using the same number of adversarial images as the original images in training and test set, the classifier achieved a test accuracy of 71\% on the adversarial images and 72.42\% on the original images. Attacking this retrained classifier, a fooling ratio of 99\% was achieved. The results are summarized in Table~\ref{deepf}.  

\begin{figure*}[h]
    \begin{center}
       \includegraphics[scale = 0.45]{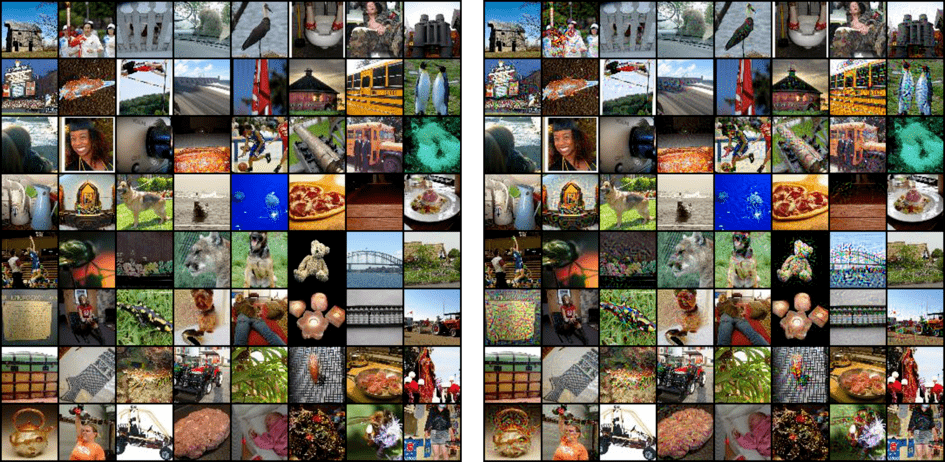}
    \end{center}
\caption{Original and adversarial images generated using DeepFool algorithm for TinyImageNet Dataset. Images on the left are the original images and the corresponding adversarial images are displayed on the right.}
    \label{tinydeep}
\end{figure*}

\begin{table*}[h]
  \centering
  \begin{tabular}{p{3cm}p{3.5cm}p{3.7cm}p{3.1cm}} 
  \toprule
    \bf Dataset     & \bf Test Accuracy on Original Images &\bf Test Accuracy on Adversarial Images & \bf Fooling Ratio after Retraining \\
    \midrule
    CIFAR-10   &\hspace{1cm}92.02\%    &\hspace{1cm}90.13\%    &\hspace{1cm}99\%    \\
    \midrule
    TinyImageNet    &\hspace{1cm}72.42\%   &\hspace{1.2cm}71\%    &\hspace{1cm}99\%  \\
    \bottomrule
  \end{tabular}
  \caption{Retraining the classifiers with the adversarial images generated using DeepFool algorithm for CIFAR-10 and TinyImageNet Dataset. The number of adversarial images used for retraining is the same as the number of original training images.}
    \label{deepf}
\end{table*}

The adversarial images along with their corresponding original images for CIFAR-10 and TinyImageNet dataset are displayed in Figure~\ref{cifar10deep} and Figure~\ref{tinydeep} respectively. It can be seen that the perturbations added to most of the images are imperceptible to the human eye. For both the datasets, the high fooling ratio even after the retraining the classifier indicates that the retraining strategy did not work as a defense in case of DeepFool attack. The reason it is successful even after retraining is because it has access to the gradients of the retrained classifier, which are used for crafting the new adversarial examples.

\subsubsection{Carlini-Wagner Attack}

We use the $L_2$ attack for our experiments because it is considered to be the strongest among the three attacks. For both the datasets, the target label is the label of the least likely class. For CIFAR-10 dataset, the initial fooling ratio is 100\%. Using the entire original data along with 50,000 adversarial images for training and 10,000 adversarial test images, test accuracy of 90.8\% is achieved on the adversarial images. Therefore, the network has learnt the images very well. After attacking the retrained classifier, the fooling ratio continues to be 99\%. Therefore, the network can be successfully fooled even after learning the adversarial images properly. 

\begin{figure}[h]
    \begin{center}
       \includegraphics[scale = 0.9]{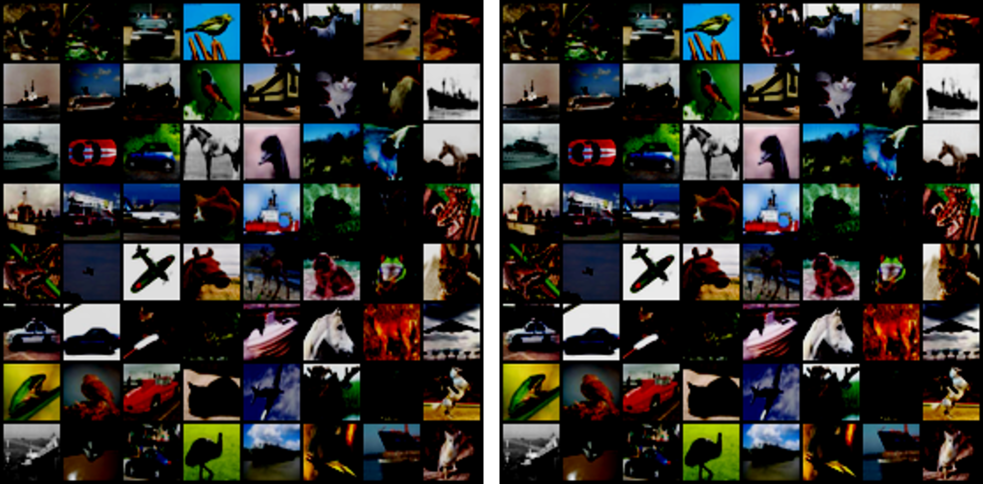}
    \end{center}
\caption{Original images and adversarial images generated using the Carlini-Wagner algorithm for CIFAR-10 Dataset. Images on the left are the original images and the corresponding adversarial images are displayed on the right. }
    \label{cifar10cw}
\end{figure}

Similarly, the initial fooling ratio is 99\% for TinyImageNet dataset. The same number of adversarial images as the original training images are used for retraining the network. The test accuracy of the network on these adversarial images is 71.3\% which is very close to the test accuracy of 72.42\% on original images. Now, the retrained network is attacked again and the fooling ratio remains the same as the initial one. The reason why it succeeds in fooling even after retraining is the same as DeepFool attack - it has access to the new gradients of the retrained network which are used for creating new adversarial images. The results for both the datasets are summarized in Table~\ref{cw}. 

\begin{table*}[h]
  \centering
  \begin{tabular}{p{3cm}p{3.5cm}p{3.7cm}p{3.1cm}} 
  \toprule
    \bf Dataset     & \bf Test Accuracy on Original Images &\bf Test Accuracy on Adversarial Images & \bf Fooling Ratio after Retraining \\
    \midrule
    CIFAR-10   &\hspace{1cm}92.37\%    &\hspace{1cm}90.80\%    &\hspace{1cm}99\%    \\
    \midrule
    TinyImageNet    &\hspace{1cm}72.11\%   &\hspace{1.2cm}70\%    &\hspace{1cm}99\%  \\
    \bottomrule
  \end{tabular}
  \caption{Retraining the network using the adversarial images generated using Carlini-Wagner algorithm for CIFAR-10 and TinyImageNet Dataset. The number of adversarial images used for retraining is the same as the number of original training images.}
    \label{cw}
\end{table*}

\begin{figure*}[h]
    \begin{center}
       \includegraphics[scale = 0.4]{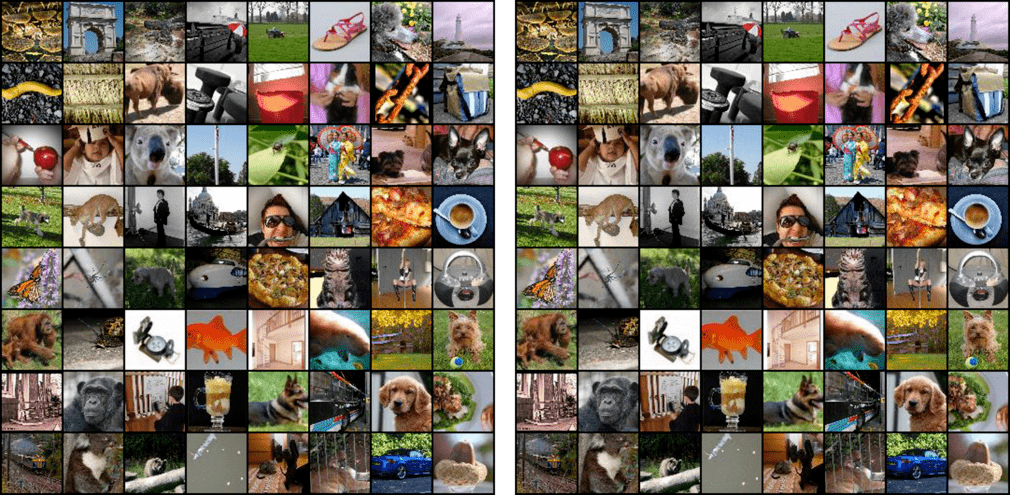}
    \end{center}
\caption{Original images and adversarial images generated using Carlini-Wagner algorithm for TinyImageNet Dataset. Images on the left are the original images and the corresponding adversarial images are displayed on the right.}
    \label{tinycw}
\end{figure*}

Figure~\ref{cifar10cw} and Figure~\ref{tinycw} show the original and adversarial images for CIFAR-10 and TinyImageNet dataset respectively. The difference between the original and adversarial images cannot be stated by merely looking at the images. Therefore, the perturbations added are invisible to the human eye. 

\subsubsection{Fast Gradient Sign Method (FGSM)}

For both the datasets, we use $\epsilon = 0.07$. For CIFAR-10 dataset, the initial fooling ratio is 92.82\%. For retraining the classifier, different number of adversarial images were used. Table~\ref{cifar10fgsm} summarizes these results.

\begin{table*}[h]
  \centering
  \begin{tabular}{p{2.8cm}p{2.2cm}p{2.4cm}p{3.1cm}}
  \toprule
    \bf Adversarial Images used for Retraining     & \bf Accuracy on Original Images     & \bf Accuracy on Adversarial Images  &\bf Fooling Ratio after Retraining  \\
    \midrule
    \hspace{0.9cm}50,000   &\hspace{0.3cm}92.45\%  &\hspace{0.35cm}90.24\%    &\hspace{0.9cm}44.72\%     \\
    \midrule
    \hspace{0.9cm}25,000     &\hspace{0.3cm}92.54\%  &\hspace{0.4cm}90.3\%   &\hspace{0.9cm}43.95\%       \\
    \midrule
    \hspace{0.9cm}12,500     &\hspace{0.3cm}92.78\%   &\hspace{0.4cm}88.5\%  &\hspace{0.9cm}46.72\% \\
    \midrule
    \hspace{1cm}9,000     &\hspace{0.3cm}92.57\%       &\hspace{0.4cm}86.1\%     &\hspace{1cm}47.8\%    \\
    \bottomrule
  \end{tabular}
  \caption{Retraining the CIFAR-10 classifier with adversarial images generated using the FGSM algorithm. The number of original training images of 50,000 remains constant and the number of adversarial images used for training is varied.}
    \label{cifar10fgsm}
\end{table*}

For TinyImageNet dataset, the initial fooling ratio is 98\%. The results for the retraining of the classifier with different number of adversarial images for TinyImageNet dataset are listed in Table~\ref{tinyimagenetfgsm}. 

\begin{table*}[h]
  \centering
  \begin{tabular}{p{2.8cm}p{2.2cm}p{2.4cm}p{3.1cm}}
  \toprule
    \bf Adversarial Images used for Retraining     & \bf Accuracy on Original Images     & \bf Accuracy on Adversarial Images  &\bf Fooling Ratio after Retraining  \\
    \midrule
    \hspace{0.8cm}100,000   &\hspace{0.4cm}72.3\%  &\hspace{0.4cm}69.2\%    &\hspace{0.8cm}48.13\%     \\
    \midrule
    \hspace{0.9cm}75,000     &\hspace{0.3cm}72.42\%  &\hspace{0.5cm}69\%   &\hspace{0.9cm}48.9\%       \\
    \midrule
    \hspace{0.9cm}50,000     &\hspace{0.3cm}72.41\%   &\hspace{0.4cm}68.1\%  &\hspace{0.9cm}49.2\% \\
    \midrule
    \hspace{0.9cm}25,000     &\hspace{0.5cm}72\%  &\hspace{0.4cm}67.1\%    &\hspace{1cm}49\%     \\
    \bottomrule
  \end{tabular}
  \caption{Retraining the TinyImageNet classifier using the original 100,000 training images and different number of adversarial images generated using the FGSM algorithm.}
  \label{tinyimagenetfgsm}
\end{table*}

\begin{figure}[h]
    \begin{center}
       \includegraphics[scale = 0.95]{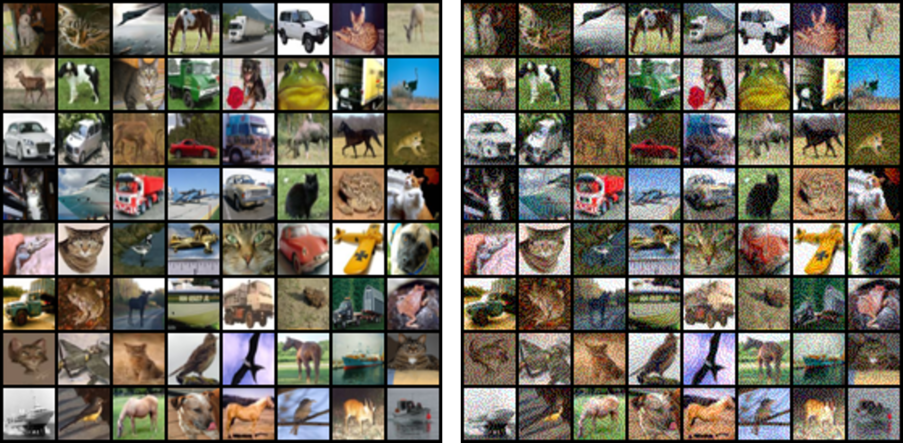}
    \end{center}
\caption{Original and adversarial images generated using the FGSM algorithm for CIFAR-10 dataset. Images on the left are the original images and the corresponding adversarial images are displayed on the right. }
    \label{cifarfgsm}
\end{figure}

\begin{figure*}[h]
    \begin{center}
       \includegraphics[scale = 0.9]{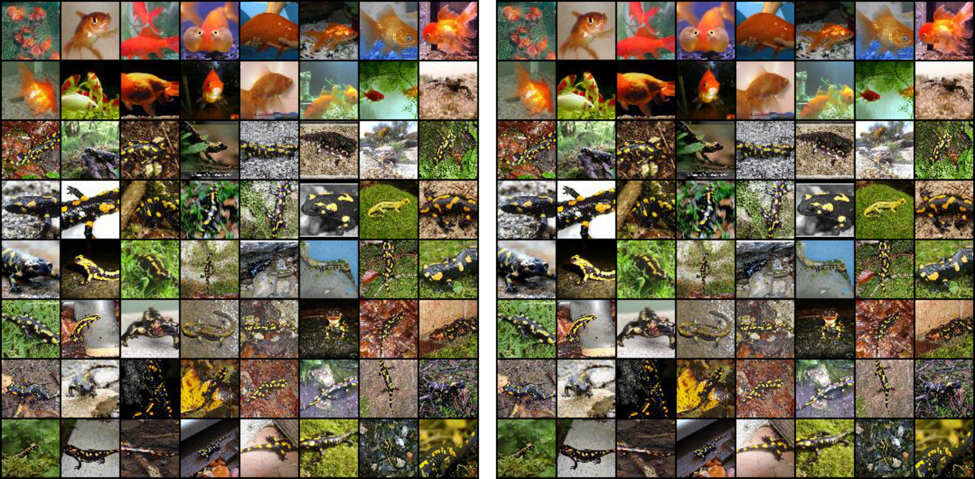}
    \end{center}
\caption{Original and adversarial images generated using FGSM algorithm for TinyImageNet Dataset. Images on the left are the original images and the corresponding adversarial images are displayed on the right.}
    \label{tinyfgsm}
\end{figure*}

Figure~\ref{cifarfgsm} and Figure~\ref{tinyfgsm} display the original and adversarial images generated using FGSM for CIFAR-10 and TinyImageNet dataset respectively. While the perturbations are visible in the CIFAR-10 images, there are completely invisible in TinyImageNet images. This is because the size of CIFAR-10 images is very small and therefore even a small change is clearly visible. 

From the results for CIFAR-10 and TinyImageNet dataset, it is clear that the network can learn the adversarial images accurately. Even though the number of adversarial images used for training decreases, the test accuracy on the adversarial images is almost as high as the test accuracy on original images. The FGSM is a white-box attack, but, the fooling ratio after retraining reduces to more than half of the initial one because it is too simple. It only adds $+\epsilon$ or $-\epsilon$ to the original image to make convert it to an adversarial one. However, the fooling ratio after retraining is not as low as the GAP attack because the new gradients of the retrained classifier are available to it which are used for generating new adversarial images. These new adversarial images can fool the retrained network to some extent indicated by a fooling ratio of approximately 45\%-49\%. 

\subsection{Producing the True Labels of the Adversarial Images}

We employ the use of the KNN algorithm to find the true labels of adversarial images generated using GAP attack. Keeping the minimum $L_2$ distance constraint on original and adversarial images in mind, we use K=1 for both CIFAR-10 and TinyImageNet dataset.  

For our experiments, the dimensions are reduced to 300 for CIFAR-10 dataset and 2000 for TinyImageNet dataset. The numbers are chosen such that the balance between the running time and the test accuracy is maintained. The accuracy of determining the labels of adversarial images is as high as 98\% on both the dataset which indicates that the labels are determined accurately. Table~\ref{knn} shows the results of these experiments. Such a high accuracy means the predicted labels for adversarial images are almost the same as their true labels. Therefore, KNN can be used for finding the true labels of the adversarial images accurately.

\begin{table*}[h]
  \centering
  \begin{tabular}{p{3cm}p{3cm}p{3.7cm}p{3cm}} 
  \toprule
    \bf Dataset     & \bf Original \hspace{1cm} Dimensions & \bf Reduced Dimensions using PCA & \bf Test Accuracy on the Labels \\
    \midrule
    CIFAR-10   & \hspace{0.3cm}3072    &\hspace{1cm}300    &\hspace{0.6cm}98.29\%    \\
    \midrule
    TinyImageNet    &\hspace{0.2cm}12288   &\hspace{0.9cm}2000    &\hspace{0.7cm}98\%  \\
    \bottomrule
  \end{tabular}
  \caption{Producing the true labels of the adversarial images generated using GAP attack for CIFAR-10 and TinyImageNet Dataset using KNN algorithm with K=1.}
    \label{knn}
\end{table*}

\section{Conclusion}

From this study, we derive a conclusion that the white-box attacks are robust to the retraining defense strategy while the black-box attacks are not. The white-box attacks like DeepFool and Carlini-Wagner can attack successfully after retraining because they have access to the new gradients of the retrained network. FGSM, despite being a white-box attack, fails to attack the retrained network with the initial fooling ratio because it is highly simple. On the other hand, the black-box attacks cannot succeed in fooling the retrained network. As having continuous access to the architecture and gradients of the network is a very strong assumption, we conclude that the neural networks can be easily defended against the adversarial attacks using retraining strategy. 

Having access to the attack algorithm to get the true labels of adversarial images is a strong assumption. Instead of relying on the attacker for the true labels of adversarial images, we employed K-Nearest Neighbor (KNN), a simple machine learning algorithm which is not specific to any particular classifier, to determine the true labels accurately. This eliminated the need for access to the attack algorithms to get the true labels of the adversarial images necessary for retraining. 

\bibliography{ref}

\end{document}